# Causal MAS: A Survey of Large Language Model Architectures for Discovery and Effect Estimation

Adib Bazgir [1], Amir Habibdoust [2], Yuwen Zhang [1*], Xing Song [3]

1. Department of Mechanical and Aerospace Engineering, University of Missouri-Columbia, Columbia, MO 65211, USA

2. Institute for Data Science and Informatics, University of Missouri-Columbia, 1 Hospital Dr. | Columbia, MO 65212.

3. Department of Biomedical Informatics, Biostatistics, and Medical Epidemiology (BBME), University of Missouri-Columbia, 1 Hospital Dr. | Columbia, MO 65212.

*Corresponding Author

## Abstract

Large Language Models (LLMs) have demonstrated remarkable capabilities in various reasoning and generation tasks. However, their proficiency in complex causal reasoning, discovery, and estimation remains an area of active development, often hindered by issues like hallucination, reliance on spurious correlations, and difficulties in handling nuanced, domain-specific, or personalized causal relationships. Multi-agent systems, leveraging the collaborative or specialized abilities of multiple LLM-based agents, are emerging as a powerful paradigm to address these limitations. This review paper explores the burgeoning field of causal multi-agent LLMs. We examine how these systems are designed to tackle different facets of causality, including causal reasoning and counterfactual analysis, causal discovery from data, and the estimation of causal effects. We delve into the diverse architectural patterns and interaction protocols employed, from pipeline-based processing and debate frameworks to simulation environments and iterative refinement loops. Furthermore, we discuss the evaluation methodologies, benchmarks, and diverse application domains where causal multi-agent LLMs are making an impact, including scientific discovery, healthcare, fact-checking, and personalized systems. Finally, we highlight the persistent challenges, open research questions, and promising future directions in this synergistic field, aiming to provide a comprehensive overview of its current state and potential trajectory.

# 1. Introduction

## 1.1 Motivation, Scope, and Definition

The integration of causal inference capabilities into artificial intelligence systems is paramount for developing robust, interpretable, and truly intelligent machines that can understand and interact with the world in a manner akin to human reasoning [14]. Large Language Models (LLMs) have shown extraordinary prowess in processing and generating human language and have made significant strides in various reasoning tasks [4, 20, 21, 29, 30]. However, their inherent mechanism, primarily based on pattern recognition and statistical correlations in vast training data, often falls short in scenarios demanding genuine causal understanding [2, 11, 26]. LLMs may generate plausible but factually incorrect or causally unsound explanations (hallucinations) [8, 11, 21, 26], struggle with out-of-distribution generalization where causal mechanisms differ from training data [9], and find it challenging to provide personalized reasoning tailored to individual contexts [27]. Multi-agent systems, wherein multiple autonomous agents interact to solve complex problems, offer a promising avenue to augment LLMs with more sophisticated causal capabilities. By decomposing complex causal tasks among specialized agents, facilitating structured debates, or enabling iterative refinement through critique and feedback, multi-agent LLM frameworks can potentially overcome the limitations of monolithic LLM approaches.

This review paper explores the emerging field of "Causal Multi-Agent LLMs." According to Figure 1, we define this area as the study, design, and application of systems where two or more LLM-based agents collaborate or interact to achieve tasks fundamentally rooted in causal inference. These tasks include, but are not limited to:

- Causal Reasoning and Counterfactuals: Agents collectively reasoning about cause-effect relationships, evaluating "what-if" scenarios, or ensuring causal consistency in their outputs.

- Causal Discovery: Agents collaborate to identify causal structures or relationships from data, potentially integrating domain knowledge or experimental interactions.

- Causal Estimation: Agents working together to quantify the strength or magnitude of causal effects, for instance, in treatment effect estimation or policy evaluation.

The scope of this review covers recent advancements that explicitly combine multi-agent paradigms with LLMs to address causal tasks. We will examine various architectural patterns, interaction protocols, evaluation methods, and application domains. The motivation is to provide a structured overview of this rapidly developing interdisciplinary field, highlighting key contributions, identifying common challenges, and outlining future research directions.

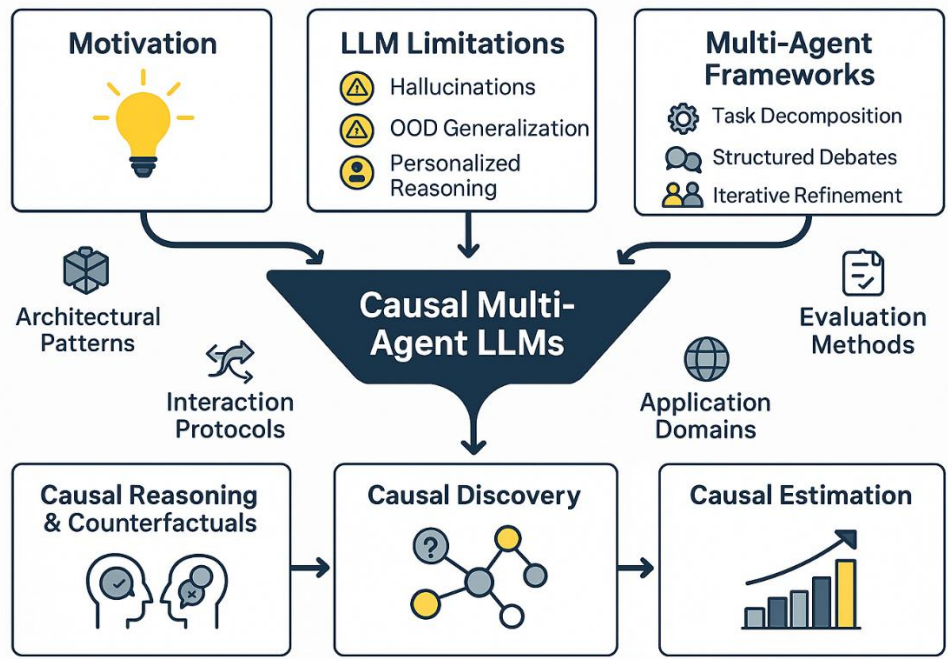

Figure 1. General workflow of causal multi-agent LLM system.

Causal AI, broadly defined as AI systems that can reason about cause and effect, holds transformative potential across numerous scientific, societal, and industrial domains. Unlike purely correlational models, causal AI aims to understand the underlying mechanisms that govern system behavior, enabling more reliable predictions, robust decision-making under interventions, and interpretable explanations [14, 15, 16].

The importance of causal AI is underscored by several factors:

- Robustness and Generalizability: Causal models are generally more robust to changes in data distribution because they capture invariant mechanisms rather than superficial statistical patterns [9, 10, 29]. This is crucial for deploying AI in dynamic or novel environments.

- Explainability and Trust: By elucidating why certain outcomes occur, causal models offer transparency, which is vital for building trust in AI systems, especially in high-stakes applications like medicine [8, 12, 30] and policy-making.

- Actionable Insights: Causal understanding allows for the prediction of outcomes under intervention (e.g., "What would happen if we implement policy X?") rather than just passive observation [12, 20, 27, 30]. This is the foundation for effective planning and decision support.

- Scientific Discovery: Uncovering causal relationships is the cornerstone of scientific progress, enabling the formulation and testing of hypotheses across disciplines from biology to economics [10, 12, 28].

- Fairness and Ethics: Causal models can help identify and mitigate biases in data and algorithms, contributing to fairer and more equitable AI systems by understanding how sensitive attributes might causally influence outcomes.

The integration of LLMs into multi-agent systems specifically for causal tasks aims to leverage the strengths of LLMs (e.g., commonsense knowledge, natural language understanding) while mitigating their weaknesses in formal causal inference through structured agent collaboration and specialized roles. This synergy is critical for advancing the frontier of causal AI and making its powerful methodologies more accessible and effective.

This review is organized as follows: Section 2 delves into how causality is treated as a primary output in multi-agent LLM systems, covering causal reasoning and counterfactuals, multi-agent causal discovery, and agentic causal estimation. Section 3 explores the diverse architectural patterns and interaction protocols that underpin these systems. Section 4 discusses evaluation methodologies, including metrics and benchmarks relevant to causal multi-agent LLMs. Section 5 surveys the wide array of application domains where these systems are being deployed. Section 6 addresses the significant challenges and open research issues in the field. Section 7 outlines promising future directions. Finally, Section 8 concludes the review with a summary of key insights and a perspective on the field's trajectory.

## 2. Causality as Primary Output in Multi-Agent AI

The pursuit of artificial intelligence that can not only predict outcomes but also understand and articulate the underlying causal mechanisms represents a significant frontier in machine learning research. Within this domain, the role of causality as a primary output, rather than a mere intermediate step, is gaining prominence. This is particularly true in complex systems where multiple intelligent agents interact. The ability of a multi-agent system to collectively infer, represent, and refine causal relationships is crucial for sophisticated reasoning, robust decision-making, and transparent explanations. This section explores methodologies where causal structures are the explicit goal of multi-agent AI systems, exploring how agents collaborate to discover, reason about, and estimate causal effects.

### 2.1 Causal Reasoning and Counterfactuals in LLM-Based Agents

Beyond discovering causal graphs, a critical aspect of causal AI is the ability of agents to engage in robust causal reasoning and to evaluate counterfactual scenarios. Causal reasoning allows agents to move beyond correlation to understand the mechanisms by which causes lead to effects, while counterfactual reasoning involves considering "what if" scenarios—how outcomes might change if certain conditions were different. These capabilities are vital for deep understanding, planning, and decision-making, especially in complex, dynamic environments. LLM-based agents, with their rich knowledge and developing reasoning skills, are increasingly being explored for these nuanced tasks.

A significant challenge in information processing tasks, such as fact-checking, is ensuring not only the accuracy of individual pieces of evidence but also the logical and causal coherence of the overall reasoning process. LLMs, despite their capabilities, can be prone to "causal errors" due to insufficient evidence or inherent hallucinations [8]. To address these issues, multi-agent systems are being designed to explicitly incorporate causal consistency checks and counterfactual evaluations. The LoCal (Logical and Causal fact-checking) framework, proposed by Ma et al. (2025) [8], exemplifies such an approach within the domain of fact-checking. LoCal employs a team of LLM-based agents to verify complex claims, with a particular emphasis on maintaining logical and causal integrity throughout the process. The system first uses a decomposing agent to break down a complex claim into simpler sub-tasks. Specialized reasoning agents then tackle these sub-tasks. The core innovation of LoCal lies in its evaluation phase, which utilizes a Logically Evaluating Agent and a Counterfactually Evaluating Agent. The Counterfactually Evaluating Agent assesses the robustness of the generated solution by challenging it with its counterfactual, determining if the solution holds when an opposite veracity prediction is assumed and checked for conflicts. This iterative refinement, guided by explicit logical and causal evaluations, aims to produce more accurate and reliable fact-checking outcomes.

The endeavor to imbue LLM-based agents with robust causal reasoning extends to ensuring the overall faithfulness and causal consistency of their generated reasoning chains, especially in knowledge-intensive tasks. The Causal-Consistency Chain-of-Thought (CaCo-CoT) framework, proposed by Tang et al. (2025) [11], is designed to bolster the faithfulness and causality of foundation models in knowledge-based reasoning through multi-agent collaboration. CaCo-CoT involves Faithful Reasoner agents that generate reasoning chains by emulating human causal reasoning and Causal Evaluator agents that scrutinize the causal consistency of these chains. The evaluators perform a "non-causal evaluation" (examining the reasoning chain against its causal flow) and a crucial "counterfactual evaluation" (testing robustness against alternative premises/answers). This multi-agent collaborative approach, which explicitly targets causal consistency, significantly improves performance on knowledge reasoning benchmarks.

The application of causal reasoning principles by LLM-based agents also enhances complex NLP tasks like machine translation, where context-dependent terms can lead to errors. CRAT (Causality-Enhanced Reflective and Retrieval-Augmented Translation), proposed by Chen et al. (2024) [13], uses a multi-agent framework where a Causality-enhanced Judge agent validates retrieved information for translation. This agent employs a "causality-driven reflection mechanism" grounded in principles of causal invariance [14, 16] and counterfactual reasoning [14, 15]. It tests if substituting a term with a potential translation preserves semantic integrity within the given context, discarding information that leads to misalignments. This ensures that the knowledge used for translation is causally consistent with the source text's intended meaning.

A particularly challenging aspect of LLM reasoning is their propensity for "hallucination" and overconfidence. Standard self-correction methods may not override an LLM's inherent biases. The CounterFactual Multi-Agent Debate (CFMAD) framework by Fang et al. (2025) [26] addresses this by compelling LLMs to explore and defend propositions that might be counterfactual to their initial biases. In an Abduction Generation stage, multiple LLM agents are assigned predetermined stances (some potentially counterfactual to the LLM's bias) and generate justifications. Subsequently, in a Counterfactual Debate stage, each abducting agent defends its stance against a skeptical critic agent. A third-party judge agent then evaluates these debates. This structured counterfactual exploration and debate helps to override initial biases and arrive at more reliable conclusions.

Further extending causal reasoning to social cognition, the ToM-agent paradigm by Yang et al. (2025) [29] equips LLM-based agents with Theory of Mind (ToM) to infer and track the unobservable mental states (Beliefs, Desires, Intentions - BDIs) of conversational counterparts. This involves agents dynamically adjusting their understanding of a counterpart's BDIs and their confidence in these inferences. A key mechanism is a counterfactual reflection method: the agent predicts a counterpart's response based on its current model of their BDIs, compares it to the actual response, and if a discrepancy exists, engages in counterfactual thinking ("What if my inferred BDIs were different?") to revise its understanding. This applies causal and counterfactual reasoning to model complex social cognitive abilities.

A crucial dimension of advancing LLM-based agents is enabling personalized causal reasoning, tailoring inferences to an individual's unique characteristics. Yang et al. (2025) [27] introduce a framework for Personalized Causal Graph Reasoning, demonstrated for dietary recommendations. An LLM agent reasons over a personal causal graph (derived from individual health data), identifies relevant causal pathways for a user's goal (e.g., glucose management), retrieves external knowledge (e.g., food databases), and then verifies recommendations by simulating dietary effects using the personal causal graph. This simulation involves estimating the causal effect of nutrient changes on health outcomes and employing counterfactual evaluation to assess if alternative choices might be more effective, thus grounding the LLM's reasoning in an individual's specific causal profile.

Finally, to bridge the gap between an LLM's general knowledge and the specific causal rules of an environment, Gkountouras et al. (2024) [20] propose a framework where an LLM agent interacts with a learned Causal World Model (CWM). The CWM is constructed using Causal Representation Learning (CRL) from environmental observations (e.g., images) and text-based actions. The LLM proposes actions in natural language; the CWM simulates the outcome in a latent causal space and returns a natural language description of the next state to the LLM. This allows the LLM agent to effectively "query" the CWM to understand the causal consequences of its actions, enhancing its planning and reasoning, especially for tasks requiring foresight over longer horizons.

## 2.2 Multi-Agent Causal Discovery

The discovery of causal relationships from data is a foundational challenge in AI. Traditional statistical causal discovery (SCD) methods, while powerful, often rely heavily on large volumes of structured data and may overlook valuable contextual information present in metadata [1]. Furthermore, the emergence of LLMs has introduced new paradigms for causal discovery by leveraging their vast knowledge and reasoning capabilities [1, 2, 3, 4, 5]. However, many initial LLM-based approaches treat the LLM as a single-agent system, which can limit reasoning capabilities when faced with complex causal relationships or large-scale, dense causal graphs [1].

To address these limitations, the concept of multi-agent causal discovery has emerged, proposing that multiple LLM-based agents can collaborate to achieve more robust and accurate causal inferences. A pioneering work in this area is the Multi-Agent Causal Discovery Framework (MAC), introduced by Le et al. (2025) [1]. MAC is presented as the first framework to explore the use of multi-agent LLMs for causal discovery, integrating both structured data analysis and metadata-driven reasoning through collaborative agent interactions. The MAC framework comprises a Debate-Coding Module (DCM) for SCD method selection and initial graph generation from structured data, and a Meta-Debate Module (MDM) where specialized LLM agents (Causal Affirmative Debater, Causal Negative Debater, Causal Judge) debate and refine the causal graph using metadata, including causal metadata derived from the DCM's output. This approach leverages the distinct roles and adversarial reasoning to enhance the robustness and interpretability of the discovered causal structures.

Beyond direct collaborative discovery, multi-agent systems can also facilitate causal discovery indirectly by serving as sophisticated data generation tools for subsequent causal analysis. Agent4Rec, a user simulator for recommendation systems by Zhang et al. (2024) [7], exemplifies this. LLM-empowered agents in Agent4Rec simulate user behaviors and preferences within a recommendation environment. The data logged from these rich, simulated interactions (e.g., movie views, ratings, user feedback) is then used as input for traditional causal discovery algorithms like DirectLiNGAM to uncover latent causal relationships within the recommendation process, such as the interplay between movie quality, popularity, exposure, and user ratings. This demonstrates how agents can generate realistic, interactional data under controlled conditions, which is then mined for causal insights.

The challenge of causal discovery in complex environments is further addressed by embodied agents that can actively interact with their surroundings to learn causal models from scratch. ADAM, an embodied causal agent for Minecraft developed by Yu and Lu (2024) [9], autonomously constructs an ever-growing causal graph (the game's technology tree) representing dependencies between items and actions. ADAM's Causal Model Module employs a two-stage process: LLM-based Causal Discovery (CD) makes causal assumptions based on interaction records (items consumed/produced by actions), and then Intervention-based CD actively experiments within the Minecraft environment (e.g., removing an item and observing the outcome of an action) to verify and refine these assumptions. While ADAM is a single agent, its methodology for building an accurate causal model through interaction and intervention is a key mechanism that could be adopted or scaled in multi-agent contexts for collaborative environmental exploration and shared causal model construction.

Further diversifying approaches, the Causal Modelling Agent (CMA) framework by Montaña-Brown et al. (2024) [10] synergizes LLM metadata-based reasoning with data-driven Deep Structural Causal Models (DSCMs). An LLM agent iteratively proposes, refines, and evaluates causal hypotheses. The process involves: Hypothesis Generation (LLM proposes an initial graph), Model Fitting (a DSCM or Deep Chain Graph Model is built based on the graph and fit to empirical data), Post-processing (LLM creates a 'memory' of changes and their impact on model fit), and Hypothesis Amendment (LLM acts as a critic to propose global and local changes to the graph). This tight loop between LLM-driven hypothesis exploration and rigorous data-driven model validation allows the CMA to effectively perform causal discovery, even in complex, multi-modal data settings like Alzheimer's Disease research.

The synergy between SCD methods and LLMs is also being enhanced by incorporating multi-modal data through sophisticated multi-agent architectures. MATMCD, by Shen et al. (2024) [28], refines an initial SCD-generated causal graph using a Data Augmentation Agent (DA-AGENT) and a Causal Constraint Agent (CC-AGENT). The DA-AGENT, using tools like web search or log lookup APIs, retrieves and summarizes relevant external textual information, creating an additional data modality. The CC-AGENT then integrates the initial graph structure with this augmented textual data, using a Knowledge LLM to explain relationships and a Constraint LLM to infer existence constraints, which are then used to refine the causal graph with the SCD algorithm. This multi-agent, multi-modal approach was shown to improve causal discovery accuracy.

These varied approaches demonstrate the versatility of multi-agent systems in causal discovery, ranging from direct collaborative graph construction and refinement, to agent-based data generation for offline analysis, to active environmental interaction and intervention for model building, and sophisticated synergy between LLM reasoning and statistical causal modeling.

## 2.3 Agentic Causal Estimation

Beyond the discovery of causal structures and qualitative reasoning about them, a critical goal of causal AI is the quantitative estimation of causal effects. This involves determining the magnitude of the impact one variable has on another, often in the context of interventions or treatments. Agentic causal estimation refers to the capability of LLM-based agents, potentially in multi-agent systems, to facilitate, automate, or even perform the complex tasks associated with estimating these causal effects from data. This can range from selecting appropriate estimation strategies and algorithms to interpreting their outputs for end-users.

The process of causal effect estimation is often intricate, demanding expertise in various methodologies to address challenges like confounding, selection bias, and heterogeneity of effects. Autonomous agents powered by LLMs are being developed to democratize access to these sophisticated techniques and to automate the analytical workflow. Causal-Copilot, an autonomous causal analysis agent proposed by Wang et al. (2025) [12], exemplifies this approach by automating the full pipeline of causal analysis, including both causal discovery and causal inference for effect estimation, for tabular and time-series data. Given a dataset and a natural language query about a causal effect, Causal-Copilot's LLM orchestrator interprets the intent, preprocesses data, and its Algorithm Selection Module chooses appropriate causal inference algorithms from an extensive library (including Double Machine Learning variants, Doubly Robust Learning methods, Instrumental Variable techniques, and matching approaches like PSM and CEM). The system then executes the algorithm, and the LLM interprets the quantitative results, generating a comprehensive report with explanations and visualizations. This makes complex causal estimation accessible to non-specialists.

Similarly, the TrialGenie framework by Li et al. (2025) [30] utilizes a multi-agent system for empowering clinical trial design through real-world evidence. A core function is performing target trial emulation (TTE) to estimate causal treatment effects from observational EHR data. Within this system, the Statistician agent plays a crucial role in causal estimation. It selects and applies methods for covariate balancing (e.g., PSM, IPTW) and outcome analysis (e.g., Cox models, Random Survival Forests) to estimate treatment effects. Other agents like the Trialist, Informatician, and Clinician collaborate to define the trial parameters, prepare data, and provide domain expertise, ensuring that the causal estimation is robust and clinically relevant. TrialGenie can also perform subgroup analyses to investigate heterogeneous treatment effects and optimize eligibility criteria based on their impact on estimated effects, showcasing a comprehensive agentic approach to causal estimation in a high-stakes domain.

These systems illustrate how LLM-based agents, particularly in multi-agent configurations, can manage the complex workflow of causal effect estimation, from understanding the research question and selecting appropriate methodologies to executing analyses and interpreting results, thereby broadening the reach and applicability of quantitative causal inference.

## 3. Architectural Patterns and Interaction Protocols

As research into causal multi-agent LLM systems matures, distinct architectural patterns and interaction protocols are beginning to emerge. These patterns define how agents are organized, how they communicate, how tasks are decomposed and allocated, and how their collective outputs are synthesized to achieve a common causal reasoning or discovery objective. The design of these architectures is crucial for enabling effective collaboration, managing complexity, and ensuring the reliability of the causal inferences drawn by the system.

One prominent architectural pattern is the pipeline framework, where tasks are sequential, and specialized agents handle discrete stages of a larger process. The Multi-Agent Pipeline Framework (MAPF) by Zhang et al. (2024) [6] for factuality evaluation via causal triple extraction is a clear example. It uses a sequence of five agents: Question Parse Agent, Search Agent, Answer Generation Agent, Fact Description Extraction Agent (extracting causal triples), and Factuality Judge Agent. Each agent's output feeds into the next, creating a structured flow for generation and verification. Similarly, Causal-Copilot [12] employs a modular pipeline orchestrated by a central LLM, encompassing User Interaction, Preprocessing, Algorithm Selection, Postprocessing, and Report Generation modules, each performing specialized functions in the causal analysis workflow. The MATMCD framework by Shen et al. (2024) [28] also follows a pipeline: an initial Causal Graph Estimator, followed by a Data Augmentation Agent (DA-AGENT) that prepares multi-modal data, which then feeds into a Causal Constraint Agent (CC-AGENT) for knowledge-driven inference, and finally a Causal Graph Refiner.

Another common pattern involves debate and collaborative refinement. The MAC framework by Le et al. (2025) [1] utilizes a Meta-Debate Module (MDM) where a Causal Affirmative Debater, a Causal Negative Debater, and a Causal Judge iteratively propose, critique, and adjudicate causal graphs. Similarly, the CFMAD framework by Fang et al. (2025) [26] for hallucination elimination employs a counterfactual debate: abducting agents generate justifications for preset stances, a critic challenges these, and the agent defends its position, with a third-party judge making the final call. CausalGPT by Tang et al. (2025) [11] uses a "reasoning-and-consensus" paradigm where Faithful Reasoner agents generate reasoning chains, and Causal Evaluator agents scrutinize their causal consistency (non-causal and counterfactual evaluation), with potential for recursive refinement until consensus.

Role-playing with iterative feedback is a more complex interaction protocol seen in the LEGO framework by He et al. (2023) [21] for causality explanation generation. Five LLMs take on distinct roles: Cause Analyst and Effect Analyst simulate bidirectional reasoning to gather information with the help of a Knowledge Master; an Explainer generates the initial explanation, and a Critic provides multi-aspect feedback, leading to iterative refinement by the Explainer. TrialGenie [30] also employs distinct roles (Supervisor, Trialist, Informatician, Clinician, Statistician) that interact dynamically beyond a simple pipeline, consulting each other and iteratively refining the clinical trial design. For instance, the Informatician might consult the Clinician on data sparsity, or the Statistician might prompt the Clinician for subgroup definitions. This system supports both a core sequential pipeline and flexible, dynamic inter-agent communication.

Agent-environment interaction architectures are crucial for embodied causality. ADAM by Yu & Lu (2024) [9] features an embodied agent with an Interaction Module (executing actions, recording observations), a Causal Model Module (LLM-based and intervention-based causal discovery), a Controller Module (planner, actor using the causal graph), and a Perception Module (MLLM for visual input). The agent learns its causal model through direct interaction and intervention in the Minecraft world. The framework by Gkountouras et al. (2024) [20] also details an agent interacting with a learned Causal World Model (CWM). The LLM agent proposes text-based actions, the CWM (built via CRL) simulates the next state in a latent causal space, and a decoder provides a natural language description of this new state back to the LLM, enabling it to "query" the causal model.

Simulation-based architectures are used when agents generate data for other processes. Agent4Rec by Zhang et al. (2024) [7] uses LLM-empowered agents to simulate user behavior in a recommendation system. These agents, with profile, memory (factual and emotional), and action modules, interact page-by-page with a recommender. The rich interaction data generated by this multi-agent simulation is then used for offline causal discovery.

Finally, some frameworks like the Causal Modelling Agent (CMA) by Montaña-Brown et al. (2024) [10] feature an LLM agent as an orchestrator in a loop with a statistical model. The LLM generates causal graph hypotheses, a DSCM/DCGM is fitted to data based on this graph, and the LLM then critiques and amends the hypothesis based on model fit and its own knowledge, using a 'memory' of past iterations. Personalized Causal Graph Reasoning by Yang et al. (2025) [27] sees an LLM agent use a pre-constructed personal causal graph to traverse paths, retrieve external knowledge, simulate dietary effects (causal estimation), and generate recommendations, guided by the structured causal information.

These patterns highlight a spectrum of complexity, from linear pipelines to dynamic, iterative, and role-based collaborations, often involving an LLM as a central reasoner, planner, or coordinator, interacting with specialized sub-agents or external tools and models. The choice of architecture and protocol is typically dictated by the specific causal task, the nature of available data, and the desired level of agent autonomy and interaction.

## 4. Evaluation & Benchmarking

Evaluating the performance of causal multi-agent LLM systems presents a unique set of challenges due to the complexity of both the causal tasks and the multi-faceted agent interactions. Standard NLP metrics may not suffice, and evaluation often needs to address the accuracy of causal claims, the coherence of reasoning, the effectiveness of collaboration, and the utility of the final output in its specific application domain.

### 4.1 Metrics for Causal Output Agents

For systems focused on causal discovery, where the output is typically a causal graph, standard graph comparison metrics are employed. These include Structural Hamming Distance (SHD),

Normalized Hamming Distance (NHD), Precision, Recall, and F1-score, which measure the differences (e.g., missing, extra, or reversed edges) between the discovered graph and a ground-truth graph [1, 7, 9, 10, 28]. For instance, MAC [1] and MATMCD [28] report these metrics on various benchmark datasets. ADAM [9] also uses SHD to evaluate the learned technology tree in Minecraft against the target graph. The Causal Modelling Agent (CMA) [10] uses NHD/BHD (Baseline Hamming Distance) ratio for comparing performance on benchmarks like Arctic Sea Ice and Sangiovese.

When agents are tasked with causal reasoning or explanation, evaluation often involves assessing the faithfulness, correctness, and coherence of the generated text. For fact-checking systems like LoCal [8], accuracy or F1-score on benchmark datasets (HOVER, FEVEROUS) is a primary metric. CausalGPT [11] also uses accuracy on knowledge reasoning benchmarks like ScienceQA and Com2Sense. For causality explanation generation, LEGO [21] uses task-specific metrics like unordered and ordered evaluation (comparing ideas and sequence in generated vs. reference explanations on WIKIWHY) and Causal Explanation Quality (CEQ) score on e-CARE, alongside human evaluation for correctness, fluency, concision, and validity.

In causal effect estimation tasks, metrics focus on the accuracy of the estimated effects. TrialGenie [30], for example, evaluates its Statistician agent by comparing estimated hazard ratios and Average Treatment Effects (ATEs) against ground truth values in synthetic datasets. For real-world emulations, it compares findings with published RCT results. Personalized Causal Graph Reasoning [27] uses a counterfactual evaluation method, calculating Mean Glucose Reduction (MGR) based on simulating the LLM's food recommendations on a ground-truth personal causal graph.

For tasks involving Theory of Mind and social reasoning, such as in ToM-agent [29], evaluation can involve human annotation to assess the similarity between inferred BDIs and true BDIs (using precision, recall, F1), and task-specific metrics like dialogue success rate (SR@t) and average turns (AT) in empathetic or persuasive dialogues.

Human evaluation plays a crucial role across many of these systems, especially for assessing the quality of generated explanations, the personalization of recommendations, or the overall coherence of agent reasoning [8, 13, 21, 27]. LLM-as-a-judge is also an emerging technique, as seen in the Personalized Causal Graph Reasoning paper [27], to assess aspects like the personalization level of reasoning.

### 4.2 Metrics for Embedded Causal Tools

Several frameworks, notably Causal-Copilot [12] and TrialGenie [30], integrate a diverse array of established causal discovery and inference algorithms as "tools" that the LLM agents can select and deploy. The performance of these embedded tools is often assessed by their ability to contribute to the overall task success. Causal-Copilot [12] conducts extensive preliminary benchmarking of over 20 individual causal discovery algorithms on synthetic datasets, varying parameters like variable size, sample size, graph density, function type, and noise distribution. Metrics like F1-score and runtime are used to create performance profiles for these algorithms, which then inform the LLM agent's algorithm selection strategy. The effectiveness of this selection process is then evaluated by Causal-Copilot's performance on compound scenarios. TrialGenie [30] evaluates its Statistician agent's ability to choose appropriate balancing and modeling methods by assessing the accuracy of the final treatment effect estimates against known values or established clinical findings. The performance of tools like RAG within the Clinician agent is implicitly evaluated by the quality and relevance of the evidence it provides for decision-making.

The primary focus is less on developing new metrics for the tools themselves, and more on the agent's intelligence in selecting the right tool (and its configuration) for the specific data and query at hand, and then correctly interpreting and utilizing its output.

### 4.3 Benchmarks and Datasets

A variety of datasets are employed to evaluate causal multi-agent LLM systems, reflecting the diverse tasks they address. For graph-recovery tasks, standard causal discovery benchmarks such as AutoMPG, DWDClimate and SachsProtein (continuous variables), as well as Asia and Child (discrete variables drawn from Bayesian networks), are commonly used, since they often come with ground-truth causal graphs or well-established structures [1, 10, 28]. When assessing causal reasoning in fact-checking or question-answering scenarios, researchers turn to benchmarks like HOVER and FEVEROUS for complex claim verification [8], ScienceQA for multimodal science question answering [11], and Com2Sense and BoolQ for commonsense reasoning [11, 26].

Specialized causal-task datasets also play an important role. WIKIWHY and e-CARE provide cause–effect pairs and natural-language explanations for causality explanation generation [21]. The MIMIC-IV electronic health record dataset is leveraged by TrialGenie to emulate clinical trials [30], and by Personalized Causal Graph Reasoning in its dietary recommendation case study [27]. AIOps datasets drawn from product-review and cloud-computing microservice systems are used by MATMCD for root-cause analysis [28].

In simulation environments designed for embodied or interactive agents, platforms like Minecraft serve as the testbed for ADAM's learning of causal game mechanics [9], while simpler custom environments such as GridWorld and iTHOR are used by the Language Agents Meet Causality work to demonstrate CWM interaction [20]. Agent4Rec initializes its recommendation-simulator agent profiles using MovieLens, Steam, and Amazon-Book datasets [7]. Finally, for evaluating social and conversational capabilities, datasets such as EmpatheticDialogue and PersuasionforGood are employed to assess Theory-of-Mind agents' performance in empathetic and persuasive dialogue settings [29].

Many works also rely on synthetic data generation to systematically evaluate specific aspects of their frameworks under controlled conditions, such as varying graph density, sample size, noise levels, or the presence of confounders [10, 12]. This allows for a more granular understanding of an agent's or system's strengths and weaknesses in different causal scenarios. The development of more comprehensive and challenging benchmarks specifically designed for causal multi-agent LLMs, covering a wider range of causal tasks and interaction complexities, remains an important area for future work.

## 5. Application Domains

As shown in Figure 2, causal multi-agent LLMs are finding applications across a diverse spectrum of domains, leveraging their enhanced reasoning and collaborative capabilities to tackle complex problems where understanding cause and effect is crucial.

One significant area is scientific discovery and research. The Causal Modelling Agent (CMA) framework, for instance, has been applied to model the clinical and radiological phenotype of Alzheimer's Disease, aiming to derive new insights into biomarker relationships [10]. Similarly, Causal-Copilot [12] is designed as an autonomous causal analysis agent to make expert-level causal analysis more accessible for scientific discovery across various fields. MATMCD [28] demonstrates its utility in AIOps by performing root cause analysis in microservice systems, a critical task for maintaining the reliability of complex IT infrastructure. The ADAM agent [9] learns the causal rules (technology tree) of the Minecraft game world, which can be seen as a form of environmental science discovery within a simulated open world.

Healthcare and Medicine represent another prominent application domain. TrialGenie [30] is a multi-agent system specifically developed to empower and accelerate clinical trial design by deriving real-world evidence from EHRs and emulating target trials for diseases like septic shock and acute heart failure. Personalized Causal Graph Reasoning, as demonstrated by Yang et al. (2025) [27], offers personalized dietary recommendations for glucose management by having an LLM agent reason over individual-specific causal graphs. The LoCal framework [8] and CausalGPT [11], while more general, are evaluated on tasks like fact-checking and knowledge reasoning which are essential for processing and verifying medical information.

Information Integrity and Fact-Checking benefit from these systems' ability to perform nuanced causal reasoning. LoCal [8] provides a multi-agent framework for logical and causal fact-checking. CausalGPT [11] aims to improve faithfulness and reduce hallucinations in knowledge reasoning. The framework by Fang et al. (2025) [26], CFMAD, uses counterfactual debating among agents to eliminate LLM hallucinations in tasks such as fact-checking and commonsense reasoning. MAPF [6] also contributes to factuality evaluation through causal triple extraction.

Personalized Systems and Recommendation are also being explored. Agent4Rec [7] creates a user simulator with LLM-empowered agents to understand user behavior in movie recommendation scenarios, which can then be used for causal discovery related to user preferences and system dynamics. Personalized Causal Graph Reasoning [27] directly provides personalized dietary advice. Conversational AI and Human-Agent Interaction are enhanced by agents with deeper causal understanding of human mental states. ToM-agent [29] empowers generative agents with Theory of Mind to simulate and track beliefs, desires, and intentions in open-domain conversations, with applications in empathetic dialogue and persuasion.

Natural Language Processing tasks that implicitly involve causality are also being addressed. CRAT [13] uses a multi-agent framework with causality-enhanced reflection for improved machine translation, particularly for context-dependent and ambiguous terms. Dr.ECI [19] infuses LLMs with causal knowledge through multi-agent decomposed reasoning for better Event Causality Identification. LEGO [21] employs a multi-agent collaborative framework for generating natural language explanations for given cause-effect pairs. Finally, the work by Gkountouras et al. (2024) [20] on bridging LLMs with Causal World Models for planning and reasoning in interactive environments points towards applications in robotics and autonomous systems where agents need to understand the causal consequences of their actions in dynamic settings.

These examples showcase the broad applicability of causal multi-agent LLMs, moving beyond traditional AI tasks to tackle problems requiring deep causal understanding, sophisticated reasoning, and collaborative problem-solving in diverse and impactful domains.

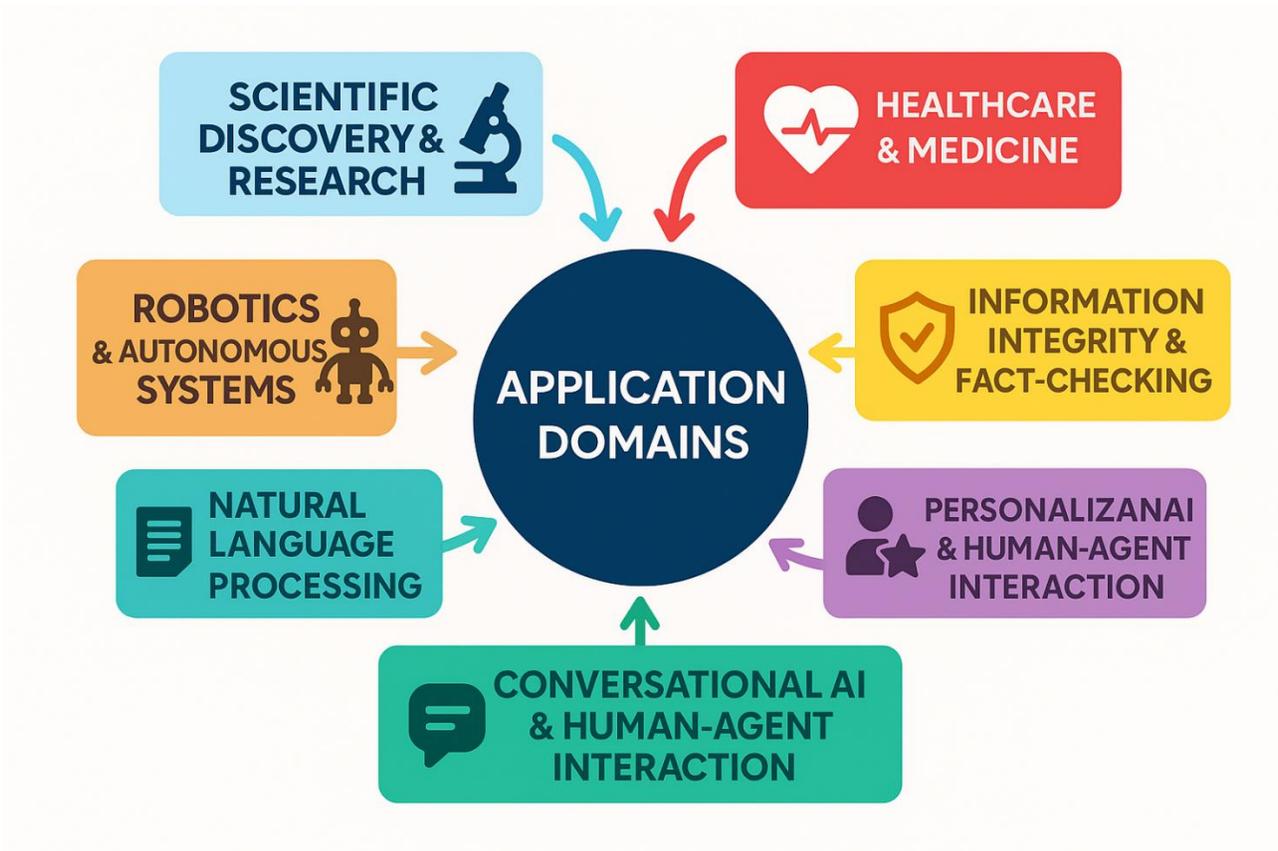

Figure 2. Application domains of causal multi-agent LLM system.

## 6. Challenges and Open Issues

Despite the promising advancements in causal multi-agent LLMs, the field faces several significant challenges and open research questions that need to be addressed to realize their full potential.

### 6.1. Reliability and Faithfulness of LLM Reasoning

A primary challenge remains the inherent tendency of LLMs to hallucinate or generate plausible but incorrect information [8, 11, 21, 26]. While multi-agent systems are often designed to mitigate this (e.g., through debate [1, 26], critique [21], or counterfactual evaluation [8, 11, 26, 29]), ensuring the causal faithfulness and logical consistency of each agent's contribution and the overall system output is non-trivial. Overcoming LLM overconfidence [26] and inherent biases [26] that may persist even in multi-agent setups is crucial.

### 6.2. Scalability and Efficiency

Many multi-agent causal frameworks involve complex interactions, multiple LLM calls, or iterative refinement loops, leading to significant computational costs and latency [1, 9, 21, 28]. For instance, edge-based or pair-based causal relationship assessments by LLMs can become intractable for large graphs [8, 28]. Developing more efficient agent communication protocols, optimized scheduling of agent tasks, and methods for more sample-efficient learning of causal knowledge are critical for scalability.

### 6.3. Knowledge Integration and Grounding

While LLMs possess vast general knowledge, effectively integrating this with domain-specific causal priors, real-time observational data, or diverse data modalities (as explored by MATMCD [28] and Causal Modelling Agents [10]) remains a challenge. Ensuring that the "causal knowledge" used or discovered by agents is accurately grounded in the specific context of the problem, rather than being based on spurious correlations from pretraining data, is essential [9, 10, 19].

### 6.4. Interpretability and Explainability

Although some frameworks aim to improve interpretability (e.g., ADAM's learned causal graph [9], LoCal's structured solutions [8], LEGO's explanations [21], Causal-Copilot's reports [12]), the reasoning processes within individual LLM agents and the emergent dynamics of multi-agent interactions can still be opaque. Developing methods to provide clear explanations for how a multi-agent system arrived at a causal conclusion or decision is vital for trust and debugging.

### 6.5. Agent Coordination and Collaboration Strategies

Designing effective collaboration protocols that go beyond simple pipelines or predefined debate structures is an ongoing research area. How to enable agents to dynamically form teams, negotiate roles, share knowledge effectively, resolve conflicts, and achieve robust consensus in complex causal tasks needs further exploration. The risk of cascading errors or "groupthink" in collaborative agent systems must also be managed [11].

### 6.6. Evaluation and Benchmarking

Standardized benchmarks and comprehensive evaluation metrics specifically designed for causal multi-agent LLM systems are still lacking [10, 12]. Current evaluations often rely on existing NLP or causal discovery benchmarks, which may not fully capture the unique capabilities or failure modes of these integrated systems. Assessing the added value of multi-agent collaboration over single-agent approaches in causal contexts requires careful experimental design.

### 6.7. Handling Complex Causal Scenarios

Real-world causal systems often involve unobserved confounders, selection bias, feedback loops, non-stationarity, and interference. While some frameworks are beginning to address aspects like unobserved confounding (e.g., CMA's Deep Chain Graph Models [10]) or heterogeneous data (e.g., Causal-Copilot [12], TrialGenie [30]), developing multi-agent LLMs that can robustly handle the full spectrum of these complexities is a major hurdle.

### 6.8. Personalization and Context Awareness

Tailoring causal reasoning to individual users or specific contexts, as explored by Personalized Causal Graph Reasoning [27] and ToM-agent [29], is a key challenge. Agents need to effectively incorporate and reason over personal data or fine-grained contextual information to provide relevant and accurate causal insights or actions.

### 6.9. Ethical Considerations

As causal multi-agent LLMs become more capable of influencing decisions in critical domains like healthcare [30] or policy, ethical considerations regarding fairness, accountability, transparency, and the potential misuse of causal inference capabilities become paramount. Ensuring that these systems operate responsibly and align with human values is an ongoing concern.

Addressing these challenges will require interdisciplinary research spanning causality, multi-agent systems, machine learning, and domain-specific expertise.

## 7. Future Directions

The intersection of causality, multi-agent systems, and LLMs is a nascent but rapidly evolving field with numerous exciting avenues for future research. Building upon the current advancements, several key directions can be identified:

### 7.1. Enhanced Causal Representation and Reasoning within Agents

***Deeper Integration of Formal Causal Models:*** Future work could focus on more deeply embedding formal causal inference mechanisms (e.g., do-calculus, counterfactual logics) within the reasoning core of individual LLM agents, enabling them to perform more rigorous causal computations beyond heuristic or pattern-based reasoning.

***Learning Causal World Models with LLM Guidance:*** Extending approaches like those of Gkountouras et al. [20] and ADAM [9], where agents learn causal models of their environment. Future systems could feature multi-agent teams collaboratively building and refining shared causal world models through diverse interactions and experiments, potentially guided by LLM-based planners or hypothesis generators.

***Handling Complex Causality:*** Developing agents that can explicitly reason about and model more complex causal phenomena such as unobserved confounding (as touched upon by Causal Modelling Agents [10]), feedback loops, non-stationarity, and causal heterogeneity will be crucial for real-world applicability.

### 7.2. Sophisticated Multi-Agent Collaboration for Causal Tasks

***Dynamic and Adaptive Collaboration Protocols:*** Moving beyond predefined roles and interaction patterns (as seen in LEGO [21] or TrialGenie [30]) towards agents that can dynamically negotiate roles, form ad-hoc teams, and adapt their collaboration strategies based on the specific causal problem and available information.

***Causal Knowledge Fusion and Conflict Resolution:*** Developing principled methods for multiple agents to share, integrate, and reconcile potentially conflicting causal beliefs or discovered causal structures to arrive at a more robust collective understanding.

***Distributed Causal Discovery and Inference:*** Exploring architectures where different agents are responsible for analyzing different parts of a large, distributed dataset or different aspects of a complex system, and then collaboratively piecing together a global causal picture.

### 7.3. Improved Evaluation and Trustworthiness

***Standardized Benchmarks and Metrics:*** The development of comprehensive benchmarks and evaluation metrics specifically tailored to assess the causal capabilities of multi-agent LLM systems across diverse tasks (discovery, reasoning, estimation) is essential [10, 12].

***Interpretability and Explainability:*** Enhancing the transparency of multi-agent causal reasoning, allowing users to understand how and why a collective causal conclusion was reached, will be key for building trust and facilitating debugging [8, 9, 12, 21].

***Robustness and Adversarial Testing:*** Systematically evaluating the robustness of these systems against noisy data, incomplete information, misleading inputs, or adversarial attacks designed to exploit weaknesses in their causal reasoning is an important research avenue.

### 7.4. Broader and Deeper Applications

***Real-World Deployment and Human-Agent Interaction:*** Moving from controlled experimental setups to real-world deployments where causal multi-agent LLMs assist human experts in domains like scientific research, healthcare decision support [27, 30], policy making, and complex system troubleshooting [28]. This will also require sophisticated human-agent interaction designs for effective collaboration.

***Ethical Frameworks for Causal AI Agents:*** Developing robust ethical guidelines and technical safeguards for deploying causal multi-agent LLMs, particularly in high-stakes decision-making contexts, to ensure fairness, accountability, and alignment with societal values.

***Lifelong and Continual Causal Learning:*** Enabling agents to continuously update and refine their causal knowledge and models as they interact with dynamic environments and new data over extended periods, as initiated by systems like ADAM [9].

### 7.5. Integration with Other AI Techniques

***Synergy with Reinforcement Learning:*** Combining causal multi-agent LLMs with reinforcement learning, where agents learn policies based on an understanding of the causal consequences of their actions, could lead to more sample-efficient and generalizable RL agents.

*Multi-modal Causal Reasoning:* Expanding capabilities to incorporate and reason over diverse data modalities (text, images, sensor data, etc.) in a causally coherent manner, building on work like MATMCD [28] and Causal Modelling Agents [10], will be critical for many applications.

The journey towards building truly causal intelligent multi-agent systems is still in its early stages, but the convergence of LLMs, multi-agent paradigms, and causal inference principles promises a future where AI can engage in deeper, more meaningful understanding and interaction with the complex causal fabric of the world.

## 8. Conclusion

The integration of causal inference principles and causal understanding with the collaborative capabilities of multi-agent systems, powered by LLMs, represents a significant and promising frontier in artificial intelligence. This review has traversed the landscape of "Causal Multi-Agent LLMs," highlighting the innovative ways in which researchers are endeavoring to imbue LLM-based agents with a deeper understanding of cause and effect. We have seen that by moving beyond the limitations of single-agent LLMs, multi-agent frameworks can tackle a diverse array of causal challenges with enhanced robustness, accuracy, and interpretability. Our exploration covered three primary facets of causality where multi-agent LLMs are making strides: causal reasoning and counterfactual analysis, where agents collaborate to ensure logical and causal consistency, evaluate hypothetical scenarios, and even model sophisticated social cognition like Theory of Mind; multi-agent causal discovery, where teams of agents work to unearth causal structures from data, leveraging debate, simulation, active intervention, or the synergy of metadata and statistical modeling, often enhanced by multi-modal information; and agentic causal estimation, where autonomous agents automate the complex pipeline of quantifying causal effects in domains like clinical trial design.

We have also examined the diverse architectural patterns and interaction protocols that enable these causal capabilities, ranging from structured pipelines and debate arenas to role-playing ensembles with iterative feedback loops and agent-environment interactions for learning world models. The evaluation of these systems, while still an evolving area, increasingly combines traditional causal metrics with task-specific assessments and human judgment, utilizing a growing set of benchmarks and real-world datasets. The application domains are already broad and impactful, spanning scientific discovery, healthcare, fact-checking, personalized recommendations, machine translation, and conversational AI. Despite the rapid progress, significant challenges and open issues persist. These include the inherent reliability and potential for hallucination in LLMs, the scalability and efficiency of complex multi-agent interactions, the effective grounding and integration of diverse knowledge sources, the need for greater interpretability, and the development of robust evaluation methodologies. Ethical considerations also loom large as these systems become more powerful.

Looking forward, the future directions are rich with potential. Enhancing the depth of formal causal reasoning within agents, designing more adaptive and dynamic collaboration strategies, creating standardized causal benchmarks for multi-agent systems, and deploying these systems in critical real-world applications while ensuring ethical alignment are key areas for continued research. The prospect of agents that can engage in lifelong causal learning and seamlessly integrate multi-modal information offers a glimpse into a future where AI can more profoundly understand and interact with the complexities of our world. In conclusion, causal multi-agent LLMs stand at a compelling confluence of several cutting-edge AI fields. By harnessing the strengths of LLMs in language and commonsense, the collaborative power of multi-agent systems, and the rigorous framework of causal inference, this domain is poised to drive significant advancements towards more intelligent, robust, and trustworthy artificial intelligence. The journey is ongoing, but the foundational work reviewed here provides a strong platform for future innovation and impact.